# Reflexive Input-Output Causality Mechanisms

Ryotaro KAYAWAKE[1], Haruto MIIDA[1], Shunsuke SANO[2], Issei ONDA[2], Kazuki ABE[2], Masahiro WATANABE[2], Josephine GALIPON[3], Riichiro TADAKUMA[3], Kenjiro TADAKUMA[2]

1. Tohoku University, Graduate School of Information Science, Sendai, Miyagi, Japan
2. Osaka University, Graduate School of Engineering Science, Toyonaka, Osaka, Japan
3. Yamagata University, Graduate School of Science and Engineering, Yonezawa, Yamagata, Japan

**Abstract**
This paper explores the concept of reflexive actuation, examining how robots may leverage both internal and external stimuli to trigger changes in the motion, performance, or physical characteristics of the robot, such as its size, shape, or configuration, and so on. These changes themselves may in turn be sequentially re-used as input to drive further adaptations. Drawing inspiration from biological systems, where reflexes are an essential component of the response to environmental changes, reflexive actuation is critical to enable robots to adapt to diverse situations and perform complex tasks. The underlying principles of reflexive actuation are analyzed, with examples provided from existing implementations such as contact-sensitive reflexive arms, physical counters, and their applications. The paper also outlines future directions and challenges for advancing this research area, emphasizing its significance in the development of adaptive, responsive robotic systems.

## 1. Introduction

Research on the implementation of reflex-based motion has a long history, with notable examples including Shigeo Hirose's pioneering work on lateral inhibition in snake robots in the early 1970's [1], and Koh Hosoda's investigation of walking robots [5]. These studies highlight an enduring interest in reflex-driven mechanisms within the field of robotics spanning several decades. The authors of this paper view soft robotics as the science of designing ultra-underactuated mechanisms, and have consistently focused their research on the fundamental mechanisms of reflex-driven systems. This research includes the development of a contact-sensitive reflex-driven omnidirectional bending arm (**Fig. 1**(i)) [8], and a reflex-driven snake-like locomotion mechanism (**Fig. 2**, **Fig. 3**) [9]. This study also introduces the concept of "intra-effector". This novel concept utilizes a combination of primary and smaller, secondary actuators to enable switching between multiple modes. An intra-effector represents a two-degree-of-freedom mechanism designed to enable the implementation of diverse characteristic or conformational changes. This mechanism allows for generating numerous functions based on the dynamic interplay between several internal states, and the forces encountered during interactions with external objects.

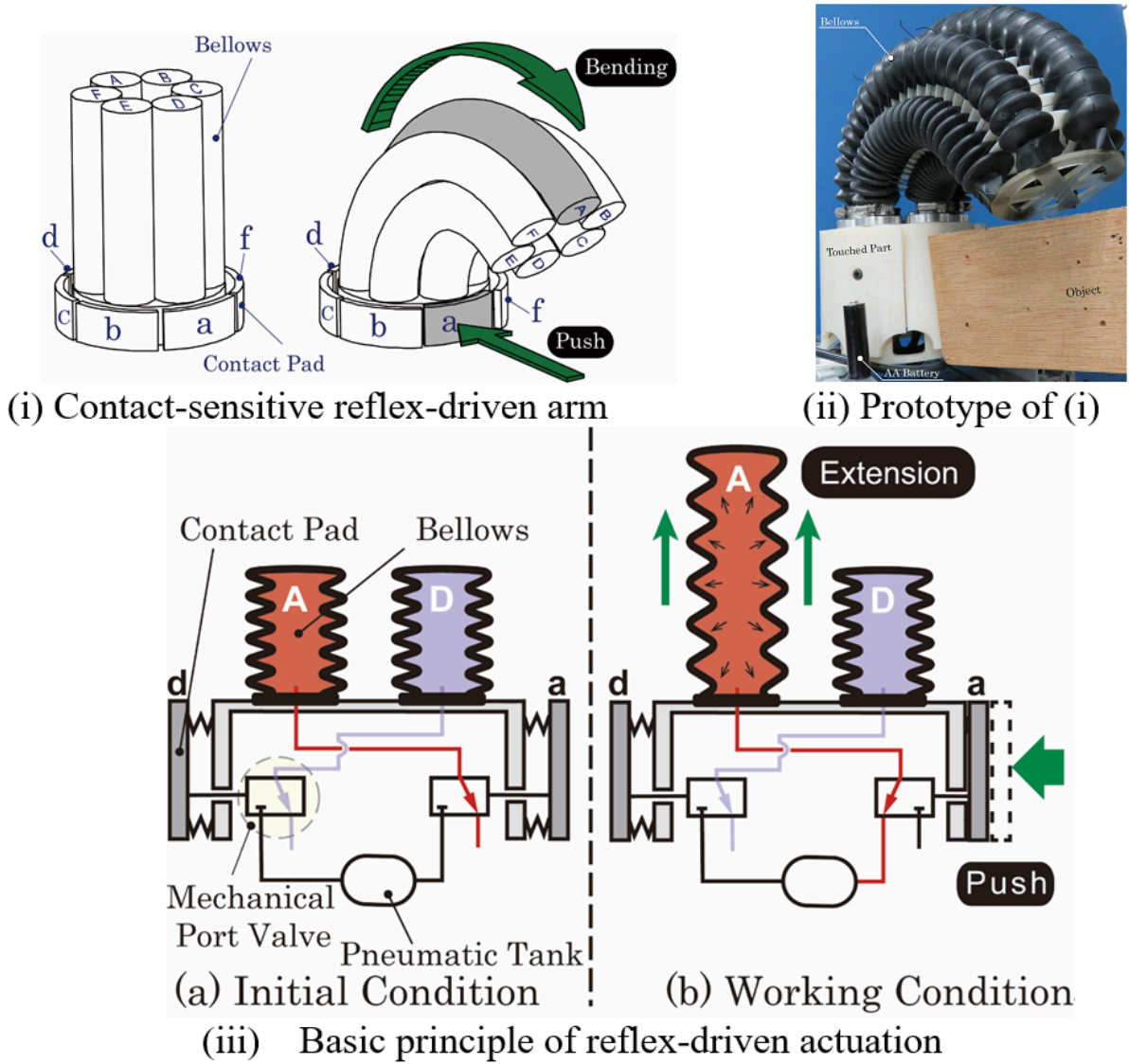

**Figure 1.** Basic principle of contact-sensitive reflex-driven mechanisms [8]

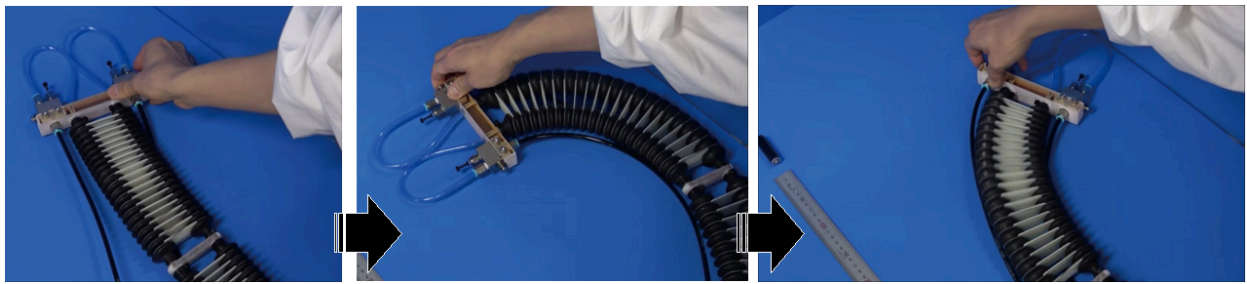

**Figure 2.** Reflex-driven snake-like robot [9]

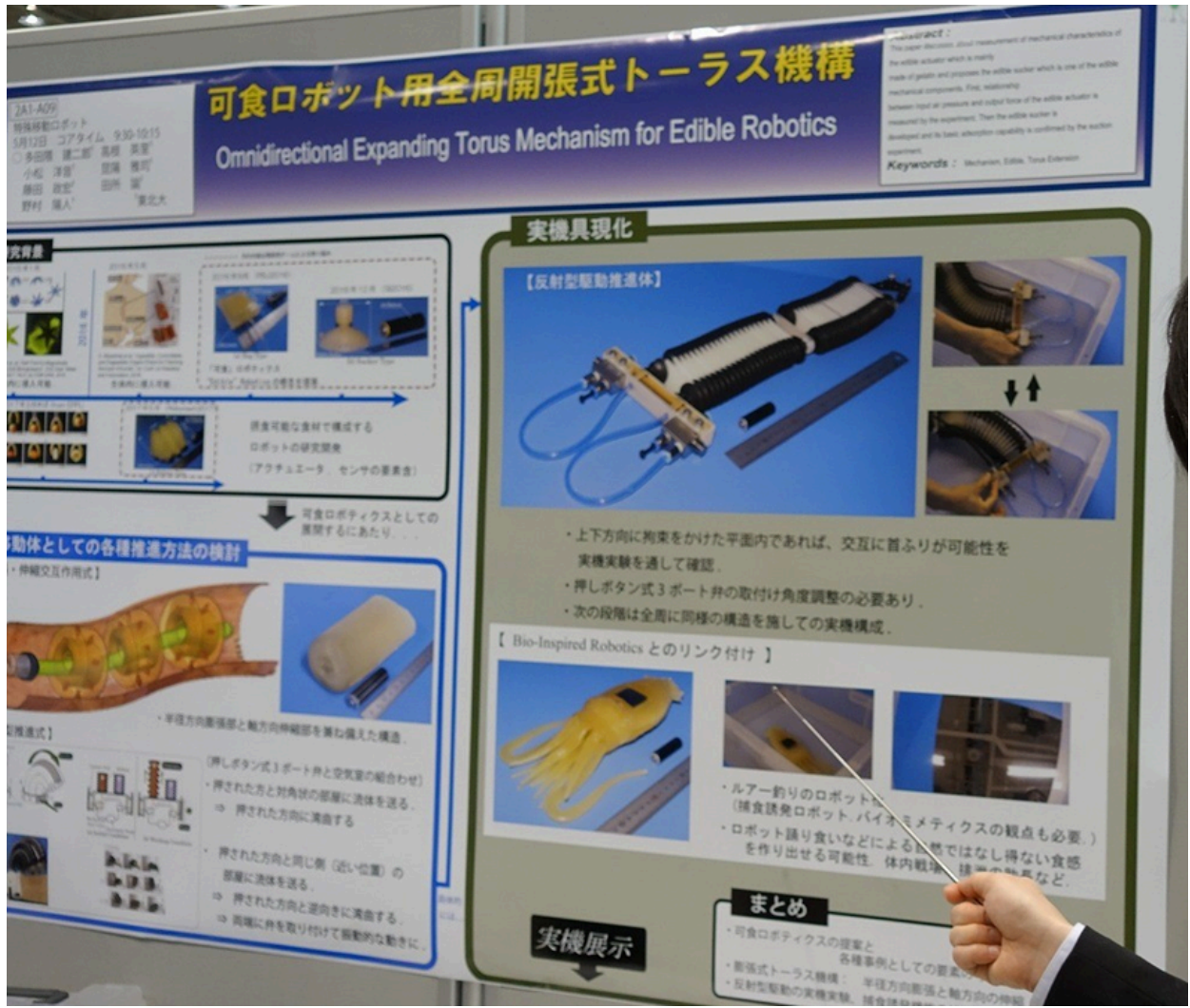

**Figure 3.** View of the poster presentation describing the reflexive snake-like robot mechanism The Robotics and Mechatronics Conference 2017 (ROBOMECH 2017) [9]

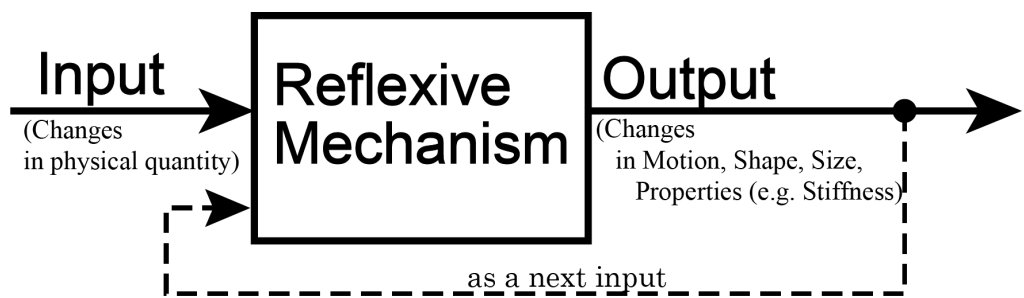

**Figure 4.** Relationship between Input and Output (SI2022 - SICE SI, 11th International Symposium on Adaptive Motion of Animals and Machines - AMAM2023, Onda *et al.* from Tadakuma Mechanism Group)

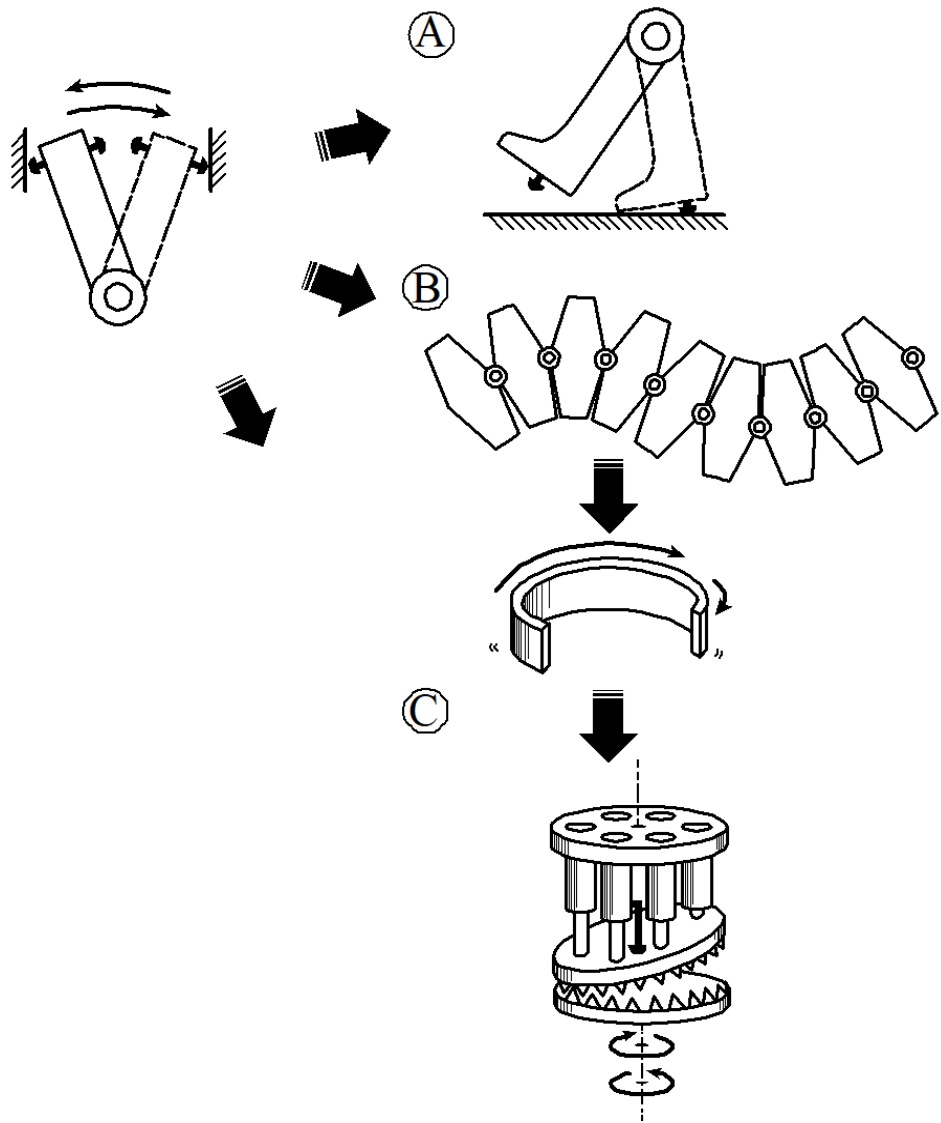

**Figure 5.** Systematic organization of reflex-driven mechanisms from the perspective of self-oscillation

## 2. Fundamental concepts

2.1 Definition of the concept of "intra-effector"

This paper introduces the concept of intra-effector within reflex-driven robotic systems. An intra-effector is a component or mechanism that facilitates mode-switching of mechanical systems, enabling a broader range of action through the following three design principles:
 ● **Functionality:** The intra-effector concept is associated with the ability of a system to switch between different modes and exhibit multiple functionalities;

● **Integration:** several actuators may be combined and designed to act together, in a coordinated or hierarchical manner, to achieve the desired mode changes and functions;
  ● **Interactivity:** intra-effectors may produce multiple functions through interactions either with external objects or through changes of internal state, or both, and are meant to facilitate interactions with the environment;

2.2 Main input/output categories and their implications

Reflex-driven systems are defined as systems where a physical quantity change of any kind in the environment acts as an input, leading to an output such as for example motion or changes in characteristics. **Fig. 4** illustrates the causality and sequentiality between input and output. Basically, there are three main categories of input that can affect such a system:

  ● **Force and Related Physical Quantities**
This category includes direct forces, as well as concepts like torque, pressure, and magnetic and electrostatic forces. These inputs act upon the system, potentially triggering a reflex-driven response. For example, the pressure of water on a system could trigger a reflex to move out of the water.

  ● **Environmental Conditions**
This category encompasses inputs such as being submerged in water, air temperature, or atmospheric pressure. Situational factors come into play, including emergency and safety considerations, threshold settings, and the configuration of trigger conditions. For instance, if the input involves entering water, the outputs may include protection against water pressure, exiting the water, or supplying oxygen. Based on this reasoning, possible output when encountering an unfavorable state or situation may involve protecting the body, safeguarding an object, escaping from the environment itself, or modifying the environment. The system's response to these conditions may also constitute a reflex mechanism. For example, a change in air temperature could trigger a system to change its shape for better heat regulation.

  ● **Output-to-Input Feedback:**
Alternatively, outputs may serve as new inputs, leading to internal, chained reactions within a system. This implies that internal changes within the system itself can also act as inputs for further reflexes. For example, an internal signal indicating low energy could trigger a reflex to seek out an energy source. This phenomenon is also observed as internally interlocked domino-effect linkages within living organisms, a concept already well-established by anatomists. Researchers in robotics, including Shigeo Hirose, have also implemented this concept.

When considering the input, it is crucial to determine whether it originates from an external (stimulus) or internal source. Factors such as the source's location, the chain of events, and the domino-effect of interlinked internal body parts must also be taken into account. Furthermore, it is necessary to determine whether the input affects internal processes (resulting in a change or switch in the characteristics of the robot itself), external processes, or both.

*2.2.1 Input/Output Characteristics*

To provide a structured overview, the this category focuses on the fundamental features of both inputs and outputs in a reflex-driven system.
 ● **Dimensions and variability:** Inputs and outputs can vary in terms of quantity, type, number of types, spatial location, distribution, time of occurrence, and temporal distribution (including multiple occurrences and vibrations).

 ● **Sources and targets:** Inputs can originate from various locations, including inside or on the body, or from external sources such as the environment, time, or specific situations. Outputs can similarly exert their effects on locations inside or on the body, or on external elements like the environment, time, or particular situations.

*2.2.2 System-Level Considerations*

● **Dynamics:** Outputs can have primary or secondary effects, differing in terms of time delay and spatial distribution.
● **System boundaries and structure:** Systems can operate in closed or open configurations. Boundaries can be eliminated, allowing for the establishment of boundaries at any desired location, and internal and external structures can be organized.
● **Extended processes:** Chemical and biological factors play a role. When considering biological factors, the five senses or responses from organs not present in humans must be considered.  Processes can occur within the body itself.

These reflex-driven actions can be associated with reflex-driven state changes, and by organizing the perspectives of internal domino-effect internal body interlocking, voluntary and involuntary movements within the body, and subsequent involuntary transitions, these actions can be further developed into a bioinspired underactuated design. This approach positions reflex-driven actions as mechanisms that enable appropriate functional adaptation, similar to the previously mentioned intra-effector and its multi-mode switching and multi-functional capabilities achieved through variable kinematic pairs.

This reflex-driven actuation can be described as a mechanism that takes changes in physical quantities as inputs and outputs them as movements or changes in characteristics. While there are methods that directly convert heat or light into displacement or force [16], we consider reflex-driven actuation to also include systems where the energy or field responsible for converting into displacement, force, or characteristic changes exists in a separate pathway, and the input serves as a trigger to connect these pathways [4]. Additionally, the changes in physical quantities at the input may arise from changes in the external environment, the internal environment of the body, or interactions between the external and internal environments. In this conversion process, it is naturally necessary to take time delays into account as much as possible. This is particularly important when using compressible fluids such as air pressure, where it is crucial to recognize and address differences in this area when realizing a physical prototype.

2.3 Generation of periodic motion through internal or body-environment coupled movements

As shown in **Fig. 2**, by internalizing the environment as a wall for the reflex-driven cord-like structure, self-excited oscillatory reciprocating motion is generated between the walls. By placing this switch near the rotation axis or, in the case of a pneumatic system, installing a switch to change the flow path at the operation's end point in a certain direction, it becomes possible to generate self-oscillatory motion. This implies that the self-excited oscillatory body can be considered as an internalized structure of reflex-driven actuation. Additionally, parallel internal coordinated motions can be seen in the leg robots by Fukuoka et al. [4] and Hosoda et al. [5], whereas linear configurations can be seen in the snake robot by Date et al. [15], and loop-shaped structures resemble the crown motor by Hirose et al. [6] (**Fig. 5**). This concept was presented at the 1st Soft Robotics Symposium on November 7-8, 2023, and SI2023, further organizing the ideas as part of the ongoing research into reflex-driven mechanisms in our lab, which began in 2015. This study has progressed from single contact to multiple contacts per unit time, and now considers aspects such as the number of contacts, duration, stroke length, force magnitude, direction, and logic. It connects all types of physical inputs with all types of physical outputs (Onda et al., AMAM2023, Tadakuma Mechanisms Group) and relates to the concepts of adaptability and underactuation.

### 3. Applications of Intra-Effector Mechanisms

As reflex-driven systems, there are various examples ranging from mechanical, fluidic, and electrical devices to chemical reactions in artificial systems, and even behaviors observed in living organisms. Examples include automatic water dispensers, *shishi-odoshi* (a traditional Japanese bamboo water fountain), siphons, gas station dispensers (utilizing the Venturi effect, though they can also be configured using a float system, making them adaptable as gravity-sensitive mechanisms), airbag helmets and other safety devices, automatic seat belt stoppers, flush toilets (where floats can be utilized as gravity-sensitive elements), fluidic components in dams, water clocks, lawn sprinklers, cup noodle timers, circuit breakers, and more.

In biological systems, examples include the stinging action of jellyfish, the mouth-closing reflex of long-tongued bees during sleep, and in nature, examples such as the tail of a cat, the inflation and display of the frill of a frilled lizard, the puffing of a pufferfish, or the courtship displays of a peacock, where changes in form occur as a response. Further output examples include the surface state changes of hedgehogs, the secretion of slime in hagfish, the ink release in octopuses and squid, the secretions of skunks and stink bugs, and the electric shocks produced by electric eels.

These examples illustrate systems where a response is output based on an input, transforming changes in various physical quantities—such as force, displacement, temperature, or light—into motion or characteristic changes (e.g., changes in shape, viscosity, elasticity, adhesiveness, friction, or surface properties). Moving forward, we will systematically organize these examples and incorporate them into internalized information branching and processing mechanisms within the body. We will also continue to advance the conceptualization and realization of practical devices, including quasi-multi-degree-of-freedom systems utilizing magnetism and information-containing mechanisms.

## 4. Conclusion

In this paper, we have further developed the concept of reflex-driven robotic mechanisms that we have been working on and have invented and realized a self-excited oscillation mechanism. Through experiments using the realized devices, we confirmed the basic effectiveness and challenges of the conceptual principles.

Moving forward, we will continue to expand the structure of these branching-process-containing robotic mechanisms systematically through the conceptualization and realization of practical devices.